%% file: main.tex
\documentclass[final,5p,times,twocolumn]{elsarticle}

\usepackage[version=3]{mhchem} 

\usepackage{graphicx}
\usepackage{amssymb}
\usepackage{scrextend}
\usepackage{xcolor}
\usepackage{amsmath}
\usepackage{enumitem}
\usepackage{pgfplots}
\usepackage{pdfpages}
\usepackage{url}
\usepackage{multirow}
\usepackage{tikz}
\usepackage{dblfloatfix}
\usepackage{array}
\usepackage{soul}
\newcolumntype{x}[1]{>{\centering\let\newline\\\arraybackslash\hspace{0pt}}p{#1}}
\usepackage{hyperref}

\usepackage{times}
\usepackage{latexsym}
\usepackage{graphicx}
\usepackage{dblfloatfix}
\usepackage{subcaption}
\usepackage{tablefootnote}
\usepackage{multicol}
\usepackage{multirow}
\usepackage{adjustbox}
\usepackage{comment}
\usepackage{placeins}
\usepackage{caption}
\usepackage{subcaption}
\usepackage{mathtools}
\usepackage{amsmath}
\usepackage{algorithm}
\usepackage{soul}
\usepackage{todonotes}
\usepackage{algpseudocode}
\algblock{Input}{EndInput}
\algnotext{EndInput}
\algblock{Parameters}{EndParameter}
\algnotext{EndParameter}

\usepackage[T1]{fontenc}

\usepackage[utf8]{inputenc}

\usepackage{microtype}

\definecolor{bblue}{HTML}{4F81BD}
\definecolor{rred}{HTML}{C0504D}
\definecolor{ggreen}{HTML}{9BBB59}
\definecolor{ppurple}{HTML}{9F4C7C}
\definecolor{deepjunglegreen}{rgb}{0.0, 0.29, 0.29}
\usepackage{color}

\journal{Computational Materials Science}

\begin{document}

\begin{frontmatter}

\title{MatSciRE: Leveraging Pointer Networks to Automate Entity and Relation Extraction for Material Science Knowledge-base Construction}

\author[label1]{Ankan Mullick\corref{cor1}}
\ead{ankanm@kgpian.iitkgp.ac.in}
\author[label1]{Akash Ghosh\corref{cor1}}
\ead{akashkgp@kgpian.iitkgp.ac.in}
\author[label1]{G Sai Chaitanya}
\ead{gajulasai@iitkgp.ac.in}
\author[label1]{Samir Ghui}
\ead{samirghui@iitkgp.ac.in}
\author[label2]{Tapas Nayak}
\ead{tnk02.05@gmail.com}
\author[label3]{Seung-Cheol Lee}
\ead{seungcheol.lee@ikst.res.in}
\author[label3]{Satadeep Bhattacharjee}
\ead{s.bhattacharjee@ikst.res.in}
\author[label1]{Pawan Goyal}
\ead{pawang@cse.iitkgp.ac.in}
\cortext[cor1]{Authors contributed equally to this work.}

\address[label1]{Indian Institute of Technology, Kharagpur}
\address[label2]{TCS Research Lab, Kolkata, India}
\address[label3]{Indo-Korea Science and Technology}

\vspace{-5mm}
\begin{abstract}
Material science literature is a rich source of factual information about various categories of entities (like materials and compositions) and various relations between these entities, such as conductivity, voltage, etc. Automatically extracting this information to generate a material science knowledge base is a challenging task. In this paper, we propose MatSciRE (Material Science Relation Extractor), a Pointer Network-based encoder-decoder framework, to jointly extract entities and relations from material science articles as a triplet ($entity1, relation, entity2$). Specifically, we target the battery materials and identify five relations to work on - conductivity, coulombic efficiency, capacity, voltage, and energy. Our proposed approach achieved a much better F1-score (0.771) than a previous attempt using ChemDataExtractor (0.716). 
The overall graphical framework of MatSciRE is shown in Fig \ref{fig:toc-graphic}. The material information is extracted from material science literature in the form of entity-relation triplets using MatSciRE.

\end{abstract}

\end{frontmatter}

\begin{figure}[!htp]
    \centering
    \includegraphics[width=8.5cm]{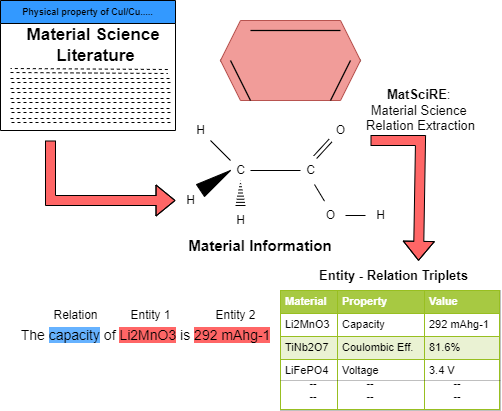}
    \caption{MatSciRE: Material Science Relation Extraction Tool}
    \label{fig:toc-graphic}
\end{figure}

\input{2introduction-rev}

\input{3RelatedWork}
\input{4Dataset}

\input{5Approach}

\input{6Experiment}

\input{7sampleop}


\section{Conclusions}
\label{sec:con}
In this work, we focus on extracting different entities and corresponding relations between two entities in the form of triplet (entity1, relation, entity2) from material science research publications. While focusing on the battery materials, we apply a distant supervision approach to create a dataset using a battery dataset previously released with the help of ChemDataExtractor \cite{huang2020database}. We apply a pointer network-based encoder-decoder model
to detect entities and relations in the form of a triplet. We also curate a gold standard annotated dataset for comparison. The extensive evaluations show that our proposed MatSciRE (Material Science Relation Extractor) performs the best in terms of F1-score to detect the triplet, outperforming ChemDataExtractor by ~6\%. Our code and dataset are publicly available, and we have deployed a web-based API where users can extract the triplets easily by uploading a manuscript.






\bibliographystyle{elsarticle-num-names}
\bibliography{main.bib}


\end{document}



\title{MatSciRE: Leveraging Pointer Networks to Automate Entity and Relation Extraction for Material Science Knowledge-base Construction}

\author[label1]{Ankan Mullick\corref{cor1}}
\ead{ankanm@kgpian.iitkgp.ac.in}
\author[label1]{Akash Ghosh\corref{cor1}}
\ead{akashkgp@gmail.com}
\author[label1]{G Sai Chaitanya}
\ead{gajulasai@iitkgp.ac.in}
\author[label1]{Samir Ghui}
\ead{samirghui@iitkgp.ac.in}
\author[label2]{Tapas Nayak}
\ead{tnk02.05@gmail.com}
\author[label3]{Seung-Cheol Lee}
\ead{seungcheol.lee@ikst.res.in}
\author[label3]{Satadeep Bhattacharjee}
\ead{s.bhattacharjee@ikst.res.in}
\author[label1]{Pawan Goyal}
\ead{pawang@cse.iitkgp.ac.in}
\cortext[cor1]{Authors contributed equally to this work.}
\cortext[cor2]{Corresponding Author}

\address[label1]{Indian Institute of Technology, Kharagpur}
\address[label2]{TCS Research Lab, Kolkata, India}
\address[label3]{Indo-Korea Science and Technology}

\section{Supplementary Information}
\label{sec:supplementary}

\subsection{Generating set of triplets}
In our dataset creation, we have used Battery database from\citet{huang2020database} consisting of a large amount of information in material science domain. The dataset is in the form of a json dump.
A sample entity from the same is shown:
\textcolor{green}
\label{Sample entity of Battery database}

\{\\
"Property"      : "Capacity", \\
"Name"          : "LiCoO2",\\
"Value"         : "130.0", \\
"Raw\_unit"      : "mAh/g", \\
"Raw\_value"     : "130", \\ 
"Unit"          : "Gram\^(-1.0)  Hour\^(1.0)  MilliAmpere\^(1.0)", \\
"Extracted\_name": "[{'Li': '1.0', 'Co': '1.0', 'O': '2.0'}]", \\ 
"DOI"           : "10.1016/j.jpowsour.2009.07.037", \\
"Specifier"     : "discharge capacity", \\
"Tag"           : "CDE", \\
"Warning"       : "None", \\
"Type"          : "None", \\
"Info"          : "{'cycle\_value': '50', 'cycle\_units': 'cycles'}", \\
"Title"         :"ELECTROCHEMICAL PERFORMANCE OF ALL SOLID STATE LITHIUM SECONDARY BATTERIES IMPROVED BY THE COATING OF LI2O-TIO2 FILMS ON LICOO2 ELECTRODE", \\
"Journal"       : "Journal of Power Sources", \\
"Date"          : "2009-07-28", \\
"Correctness"   : "T" \\
\}\\

We only extract 'property', 'name', 'value', 'raw\_unit' and 'raw\_value' for our task.

\subsection{Pre-processing of sentence triplet pairs}
After the generation of sentence triplet pairs, an important pre-processing step is required. The set of generated sentence-triple pairs so far do not have the positions of the entities and relations in the sentence. This dataset needs to be converted to a structured format and also needs to store the starting and ending indexes of the entities and the relation along with the sentence triplet pair so that it can be used in the pointer-network model.  
A single entry in the unstructured dataset, consists of a sentence along with a triplet and a document id. We convert this entry to a json object with `id', `docId' and index details of entities and relations. This forms the 'structured dataset'. A sample entry in json format is shown:
\\\newline
\{ \\
"id": 5507, \\
"docId": 2011, \\ 
"sentText": "The voltage plateau at around 2.0 and 1.7 V are verified the lithium ion insertion / extraction of anatase TiO2 .", \\ 
"relationMentions": [ \\
\hspace*{1cm}    \{"arg1Text": "TiO2", \\ 
\hspace*{1cm}     "arg1StartIndex": 19, \\
\hspace*{1cm}     "arg1EndIndex": 19, \\
\hspace*{1cm}     "relText": "Voltage", \\
\hspace*{1cm}     "relStartIndex": 1, \\
\hspace*{1cm}     "relEndIndex": 1, \\
\hspace*{1cm}     "arg2Text": "1.7 V", \\
\hspace*{1cm}     "arg2OriginalText": "1.7 Volt\^(1.0)", \\ 
\hspace*{1cm}     "arg2StartIndex": 7, \\
\hspace*{1cm}     "arg2EndIndex": 8\} \\
 ], \\
"numTriples": 1 \\
\}

The \textit{sentText} field contains the sentence that is the text, \textit{id} is just a unique identifier for each sentence. \textit{relationMentions} field contains a list whose size is equal to the number of triplets generated using the sentence pointed to by the \textit{sentText} field of the same entry. Each element in this list corresponds to a triplet, and the fields \textit{arg1Text}, \textit{relText}, \textit{arg2Text} denote the two entities and the relation parts of the triplet as they appear in the sentence. It must be noted, that the second entity part of the triplet may not necessarily exactly match it's occurence in the text. Hence, \textit{arg2OriginalText} field contains the second entity as it appears in the triplet. \textit{arg1StartIndex} and  \textit{arg1EndIndex} are word-based indices starting with 0 and denote the positions where the first entity starts and ends in the sentence text. \textit{arg2StartIndex}, \textit{arg2EndIndex} and \textit{relStartIndex}, \textit{relEndIndex} denote the same for second entity and the relation parts of the triple.

\subsection{Annotation}

 We selected three annotators fulfilling several criteria: The annotators should have strong domain knowledge, along with a good proficiency in English. The annotators are experienced in similar tasks (with earlier publications and public datasets), hence they are considered reliable for the annotation task. The initial labelling is done by two annotators and annotation conflicts are checked and resolved by the third annotator after discussion with the first two annotators. So, the final judgment is based on the three annotators’ discussion and conflict resolution. One common conflict arose regarding the inclusion of the material name. For example, in the following sentence -``Tin phosphide ( Sn4P3 ) has been a promising anode material for SIBs owing to its theoretical capacity of 1132 mA h g 1 and high electrical conductivity of 30.7 S cm 1 " - One annotation contained "Tin phosphide ( Sn4P3 )" and another contained only "Sn4P3". Finally, "Tin Phosphide" was not included. In another scenario - ``For the NG / S–40TiO2 electrode showing slightly inferior performance, the reason could be that TiO2 has intrinsically low Li-ion diffusivity ( ${10}^{-12}$ to ${10}^{-9}$ S $cm{-1}$] ) and electronic conductivity (${10}^{-12}$ to ${10}^{-7}$ S $cm{-1}$), which could offset the positive effect." - One annotation was - `[(TiO2, Conductivity, $10^{-12}$ S $cm{-1}$), (TiO2, Conductivity, ${10}^{-7}$ S $cm{-1}$)]' (2 different triplets) and another was - `[TiO2, Conductivity, ${10}^{-7}$ to $10^{-7}$ S $cm{-1}$]' (one triplet). Here, the conflict was based over the number of triplets, whether to annotate with one triplet, or two triplets. Finally, the decision was made to annotate the instance with two triplets. The above examples show the challenge in annotating the dataset, where annotators with deep domain knowledge disagree in some cases. We reported the inter-annotator-agreement including the conflicts so the reported value was 0.82 - which is almost perfect $\kappa$ score as described in (Landis and Koch 1977).

\subsection{Experimental Results}

\begin{table*}[]
\centering
\begin{adjustbox}{width=1\linewidth}
\begin{tabular}{|c|c|c|c|c|c|c|}
\hline
 & \multicolumn{3}{c|}{\textbf{Distantly supervised dataset}} & \multicolumn{3}{c|}{\textbf{Annotated ground truth dataset}} \\\hline
 & \textbf{Pr, $sd$} & \textbf{Re, $sd$} & \textbf{F1, $sd$} &  \textbf{Pr, $sd$} & \textbf{Re, $sd$} & \textbf{F1, $sd$} \\ \hline
\textit{\textit{Voltage}} & 0.92,0.012 &
0.88,0.008 &
0.899,0.002 & 0.943,0.005 &
0.838,0.017 & 0.912,0.008 \\ \hline
\textit{\textit{Capacity}} & 0.921,0.029 & 0.896,0.007 &
0.909,0.001  & 0.95,0.006 & 0.82,0.015 &
0.885,0.004  \\ \hline
\textit{\textit{Conductivity}} & 0.809,0.018 &
0.717,0.037 &
0.76,0.009 & 0.849,0.008 & 0.849,0.005 &
0.849,0.018  \\ \hline
\textit{\textit{Coulombic Efficiency}} & 0.926,0.003 & 0.904,0.021 & 0.915,0.005 & 0.923,0.011 &
0.93,0.003 &
0.926,0.004   \\ \hline
\textit{\textit{Energy}} &  0.928,0.002 & 0.895,0.019 & 0.911,0.004 & 0.909,0.004 & 0.937,0.005 &
0.923,0.020  \\ \hline
\textbf{macro score} & 0.921,0.024 &
0.883,0.003 &
\textbf{0.905},0.014  &  0.926,0.003 &
0.86,0.015 & \textbf{0.901},0.006   \\ \hline
\end{tabular}
\end{adjustbox}
    \caption{Precision (Pr), Recall (Re), F1-Score (F1) and their respective standard deviations ($sd$) of PNM (BERT) on distantly supervised dataset and annotated ground truth dataset}
  \label{tab:bert-all}

\end{table*}

PNM(MatBERT) performs the best among all the models. We show the performance of PNM(BERT) on both distantly supervised and manually annotated datasets in table ~\ref{tab:bert-all}. We compare the results for PNM (BERT) model with PNM (MatBERT) on all relations for distantly supervised dataset and manually annotated dataset in table ~\ref{tab:bert-matbert-all} and table ~\ref{tab:bert-mat-annnotation} respectively. We observe that, in both the distantly supervised dataset and the annotated ground truth dataset, PNM (BERT) has better precision than PNM (MatBERT). This may be due the fact that PNM (BERT) predicts less number of triplets. But PNM (MatBERT) recall is far superior which is why the overall F1-score is higher for PNM (MatBERT).

\begin{table*}[]
\centering
\begin{adjustbox}{width=0.9\linewidth}
\begin{tabular}{|c|c|c|c|c|c|c|}
\hline
 & \multicolumn{3}{c|}{\textbf{PNM (Bert)}} & \multicolumn{3}{c|}{\textbf{PNM (MatBERT)}} \\\hline
 & \textbf{Pr, $sd$} & \textbf{Re, $sd$} & \textbf{F1, $sd$} &  \textbf{Pr, $sd$} & \textbf{Re, $sd$} & \textbf{F1, $sd$} \\ \hline
\textit{\textit{Voltage}} & 0.92, 0.012 &
0.88, 0.008 &
0.899, 0.002 & 0.918, 0.006 &
0.893, 0.005 & 0.906, 0.005 \\ \hline
\textit{\textit{Capacity}} & 0.921,0.029 & 0.896,0.007 &
0.909,0.001  & 0.914, 0.029 & 0.901, 0.004 &
0.907, 0.015  \\ \hline
\textit{\textit{Conductivity}} & 0.809,0.018 &
0.717,0.037 &
0.76,0.009 & 0.875, 0.007 & 0.859, 0.009 &
0.866, 0.002  \\ \hline
\textit{\textit{Coulombic Efficiency}} & 0.926,0.003 & 0.904,0.021 & 0.915,0.005 & 0.925, 0.0085 &
0.919, 0.003 &
0.921,0.006   \\ \hline
\textit{\textit{Energy}} &  0.928,0.002 & 0.895,0.019 & 0.911,0.004 & 0.905, 0.006 & 0.942, 0.012 &
0.923, 0.009  \\ \hline
\textbf{macro score} & 0.921,0.024 &
0.883,0.003 &
0.905,0.014  &  0.92, 0.008 &
0.907, 0.004 & \textbf{0.913}, 0.002   \\ \hline
\end{tabular}
\end{adjustbox}
    \caption{Precision (Pr), Recall (Re), F1-Score (F1) and their respective standard deviations ($sd$) of PNM (BERT) and PNM (MatBERT) on distantly supervised corpus}
  \label{tab:bert-matbert-all}

\end{table*}

\begin{table*}[]
\centering
\begin{adjustbox}{width=0.9\linewidth}
\begin{tabular}{|c|c|c|c|c|c|c|}
\hline
 & \multicolumn{3}{c|}{\textbf{PNM (BERT)}} & \multicolumn{3}{c|}{\textbf{PNM (MatBERT)}} \\\hline
& \textbf{Pr, $sd$} & \textbf{Re, $sd$} & \textbf{F1, $sd$} &  \textbf{Pr, $sd$} & \textbf{Re, $sd$} & \textbf{F1, $sd$} \\ \hline

\textit{\textit{Voltage}} & 0.943,0.005 &
0.838,0.017 &
0.912,0.008 &
0.922,0.003 &
0.898,0.013 &
0.912,0.008 \\ \hline
\textit{\textit{Capacity}} & 0.95,0.006 &
0.82,0.015 &
0.885,0.004 &
0.885,0.008 &
0.901,0.002 &
0.893,0.012 \\ \hline
\textit{\textit{Conductivity}} & 0.849,0.008 &
0.849,0.005 &
0.849,0.018 &
0.868,0.034 &
0.868,0.007 &
0.868,0.004 \\ \hline
\textit{\textit{Coulombic Efficiency}} & 0.923,0.011 &
0.93,0.003 &
0.926,0.004 &
0.933,0.006 &
0.922,0.003 &
0.927,0.005 \\ \hline
\textit{\textit{Energy}} & 0.909,0.004 &
0.937,0.005 &
0.923,0.020 &
0.911,0.015 &
0.953,0.006 &
0.932,0.008 \\ \hline
\textbf{macro score} & 0.926,0.003 &
0.86,0.015 &
0.901,0.006 & 0.919,0.005 &
0.911,0.004 &
\textbf{0.915},0.007\\ \hline
\end{tabular}
\end{adjustbox}
    \caption{Precision (Pr), Recall (Re), F1-Score (F1) and their respective standard deviations ($sd$) of PNM (BERT) and PNM (MatBERT) on annotated ground truth datasets}
  \label{tab:bert-mat-annnotation}

\end{table*}

\bibliographystyle{elsarticle-num-names}
\bibliography{sample-v1.bib}

%% file: 2introduction-rev.tex
\section{Introduction}
\label{S:1}

The material science domain possesses numerous information. This knowledge must be extracted from material science datasets to improve or discover the latest materials. Current big data and machine learning-based approaches can help bridge the gap between theoretical concepts and current
material infrastructure limitations. People have worked on different aspects of material science literature in the last decade. One of
the critical directions of material science research is the battery database which is essential in the modern energy system.

Batteries are made up of complex material systems \cite{nitta2015li}. Proper analysis of the battery database would lead us to discover new materials. 
The battery is a crucial element of any electrical device with various utilisations. Researchers focus on building high-capacity, efficient, safe batteries with various industrial applications. Relations must be retrieved from unorganized data sources to build a concrete battery knowledge database. Research papers describing battery materials can be a potential source from which these relations can be obtained. Hence, an automated relation extraction method from material science research articles would be a great leap towards overcoming present constraints and achieving desired outcomes. 

For example, Table \ref{tab:intro-example} shows two entities (`Na0.35MnO2' and `42.6 Wh kg 1') and corresponding relation (`Energy') extraction from an input sentence ``The energy density based on AC and nanowire Na0.35MnO2 is 42.6 Wh kg 1 at a power density of 129.8 W kg 1.'' of a research article.

\begin{table*}[!htb]
\centering
\caption{Battery Database Entity and Relation}
\label{tab:intro-example}
\captionsetup{justification=centering,margin=0.0mm}
\begin{adjustbox}{width=0.85\linewidth}
\begin{tabular}{|c|c|c|c|}
\hline
 \textbf{Sentence} & \textbf{Entity 1} & \textbf{Entity 2} & \textbf{Relation}   \\ \hline

The energy density based on AC and nanowire Na0.35MnO2 & Na0.35MnO2 & 42.6 Wh kg 1 & Energy\\
is 42.6 Wh kg 1 at a power density of 129.8 W kg 1. & & &\\\hline

\end{tabular}
\end{adjustbox}
\end{table*}

Recent Natural Language Processing (NLP) and Deep Neural Networks (DNN) techniques can facilitate relation extraction and automatically build the battery database. These methods can detect various entities in the research articles discussing battery materials and predict relations between two entities.

To use these NLP and DL techniques, we first obtain an annotated dataset containing the sentences and the corresponding entity-relation triplets\footnote{A triplet consists of two entities connected with a relation.}. We use a distant supervision approach on top of the battery dataset released by \citet{huang2020database} 
to develop a pseudo-labeled dataset. Human curators annotate a random sample from this dataset to prepare a gold standard dataset for evaluation. We train a pointer-network model (PNM), adapted from  \citet{nayak2020effective} with MatBERT embeddings \cite{walker2021impact} for extracting entity-relation triples from material science papers.  
While evaluating the same test set as used by ChemDataExtractor, we find that MatSciRE outperforms ChemDataExtractor by 6\% (F1-score). 

%% file: 3RelatedWork.tex
\section{Related Work}
\label{sec:related}

To give a broad overview of the existing works in this domain, we categorize various earlier works into two categories - Material Science Information Extraction and Relation Extraction frameworks. 

\subsection{Material Science Information Extraction:} \citet{vahe2019unsupervised} train the skip-gram variant of word2vec~\cite{mikolov2013efficient} model over a large corpus of material science research papers and demonstrate that the trained embeddings capture complex material science concepts like the structure-property relationships in materials.

Wikipedia holds much information on various subjects and domains in the form of unstructured text, images, and tables. DBPedia~\cite{auer2007dbpedia} is a community effort that aims to extract structured data from the unstructured Wikipedia and make the extracted data available to the web.  
DBPedia also allows other external datasets to be connected to its datasets. This helps DBPedia to expand its knowledge beyond just wikipedia-based datasets. The triplets that constitute DBPedia's datasets are available publicly. 
Some studies aim at constructing knowledge graphs using DBPedia dumps and other structured datasets for the material science domain. \citet{zhang2014keam} design a system to recommend metal materials to material scientists based on the triplets made available by DBPedia using semantic distance measurement.
\citet{zhang2017mmkg} propose an approach to build a Knowledge Graph for metallic materials using DBpedia and Wikipedia. 

\citet{weston2019named} train a Named Entity Recognition (NER) model to identify summary-level information from materials science documents such as the name of inorganic materials, sample descriptors, etc. The authors train a BiLSTM-CRF model to classify tokens. \citet{guha2021matscie} present MatSciIE, a tool for materials science information extraction like - the name of materials, code, parameters, etc. They build a Bi-LSTM-CRF-Elmo model for entity retrieval. \citet{mullick2022using} build sentence level classification for entity extraction from material science. \ \citet{luan2018multi} retrieve entities and relations between multiple entities on the SCIERC dataset. Five hundred scientific articles from various domains are processed, and abstracts are used to create the SCIERC dataset. 
 \citet{beltagy2019scibert} which pretrains a BERT model on scientific articles of biomedical documents to improve the effectiveness of different scientific NLP tasks. Researchers also built BERT based domain specific pre-trained models on material science (MatBERT ~\cite{walker2021impact} and MatSciBERT~\cite{gupta2022matscibert}) and battery materials (BatteryBERT~\cite{huang2022batterybert}). We use both of the models as baselines.

\subsection{Relation and Entity Extraction:} The relation extraction task has been well explored in the field of NLP. 
\citet{mintz2009distant} explored distant supervision based relation extraction task using Freebase~\cite{bollacker2008freebase}, avoiding the domain dependence of automatic content extraction task. \citet{riedel2010modeling} propose a distant supervision-based relation extraction task to extract various entities without using training annotation data. 
\citet{hoffmann2011knowledge} work on a weakly supervised method to develop multi-instance learning to retrieve overlapping relations. \citet{zeng2015distant} apply deep neural network models to extract lexical and sentence level features for relation classification and propose CNN (convolutional neural network) based distant supervision technique for relation extraction. \citet{shen2016attention} build a novel attention-based CNN framework for relation classification. \citet{jat2018improving} suggest combining bi-directional word-based and entity-centric attention models to improve distant supervision-based relation extraction task. RESIDE \cite{vashishth2018reside} is a Distantly-supervised Relation Extraction (RE) method that employs Graph Convolution Networks
to encode text. \citet{ye2019distant} develop a distantly supervised relation extraction method that can handle noisy data. \citet{guo2019attention} propose attention-based graph convolutional networks that can extract various relations from the full dependency tree as input. \cite{mullick2022framework, mullick2023intent, mullick2022fine, mullick2023novel, mullickexploring} work on intent and entity extraction. \cite{mullick2017generic, mullick2016graphical, mullick2018identifying, mullick2018harnessing, mullick2019d,mullick2017extracting} aim at opinion-fact entity extraction. \citet{nayak2020effective} jointly extract entities and classify relations between entities using a pointer network-based model. \citet{huang2020database} generate a set of data records from a corpus of materials science research papers. Unlike others, the authors work directly with the unstructured text using a popular NLP toolkit called `ChemDataExtractor'. Authors, however, modify ChemDataExtractor 1.5 to perform better on the specific domain of battery materials. ChemDataExtractor is then used to extract a dataset of 292,313 records from a corpus consisting of 229,061 research papers. Each extracted data record contains an entity (usually the chemical formula of a material), a value with a unit, and a relation relating the entity with the value, among other things (We use this approach as a baseline). 

A group of researchers also focused on the joint extraction of entities and the corresponding relations among them. CoType \cite{ren2017cotype}, \citet{miwa2016end}, \citet{bekoulis2018joint}, novel tagging scheme by \citet{zheng2017joint} and BiLSTM-CRF-based model by \citet{nguyen2019end} are end-to-end models to jointly extract relation and entities together. \citet{katiyar2016investigating} investigate jointly extract opinionated entities and the relation between two entities. 

However, in our particular problem, there can be multiple triplets in the same sentence, and the entities within the triplets may also overlap. This creates a bottleneck for neural sequence tagging-based joint entity extraction models \cite{zheng2017joint}. 

We use an encoder-decoder framework to address the issue of relation extraction. Encoder-decoder framework has been utilized in various natural language processing problems like - knowledge base creation using n-gram attention mechanism \cite{distiawan2018gtr}, text generation from structured data \cite{marcheggiani2018deep}, cross lingual information extraction \cite{zhang2017mt}, open information extraction \cite{cui2018neural}, machine translation \cite{bahdanau2014neural,luong2015effective}, pointer networks \cite{vinyals2015pointer} based relation extraction \cite{nayak2020effective}, question answering \cite{kundu2018question}, etc. 
Our encoder-decoder framework is a pointer network-based architecture to overcome earlier limitations which retrieves entities and the corresponding relations in the form of triplets from sentences in material science literature.

%% file: 4Dataset.tex
\section{Dataset}
\label{sec:dataset}

\citet{huang2020database} develop the database of battery materials with entities and corresponding relations using a rule-based phrase parsing method, resulting in a database consisting of 292,313 data records from 229,061 academic papers. They suggest 17,354 unique chemicals and up to five material properties: capacity, voltage, conductivity, Coulombic efficiency and energy. The rule-based approach to specific entities helps to identify the relations. This database only contains relations and entities without any sentence-level mapping. 

However, training a system to extract entities and relations from an article also requires corresponding sentence information. Therefore, to train an end-to-end framework for extracting entities and relations from material science articles, we need a dataset where sentences of an article are annotated with the triplet information. So, at first, on top of the dataset by \citet{huang2020database}, we use the distant supervision technique by \citet{mintz2009distant} to build the entity-relation triplet pairs and corresponding sentences from material science articles. 
The triplet is of the form ($entity1, relation, entity2$). For example, in the manuscript by \citet{yu2016atomic}, the extracted triplet from the sentence, ``The low voltage profiles (average of around 2.2 V) of Li–S batteries are well compensated by their high energy densities''  is (Li-S, voltage, 2.2 V). Later, we also manually annotate a random sample of this dataset to create a {\sl gold standard dataset} for evaluation. We use a large part of the {\sl distantly supervised dataset} to train the model, another part of the data is utilized to tune various model parameters and hyper-parameters, and the rest of the dataset is considered as blind test data to evaluate the model performance. We call this the {\sl distantly supervised dataset.} 

The detailed steps of the dataset creation process are discussed next.

\subsection{Set of Triplets}

The first step towards creating the dataset is to obtain a set of triplets generated over material science domain. While not many contributions have been made in this direction in the materials science domain to the best of our knowledge, we found the work of \citet{huang2020database} to be relevant. The authors generate a database of records for the battery materials domain using a large corpus of materials science research papers. A sample entry from the JSON dump is shown in the Supplementary Material.

Each instance consists of "Property", "Name", "Value" and "Unit" among other fields. "Raw\_value" and "Raw\_unit" store the simplified value and unit, respectively, which is useful for text processing. "Extracted\_name" stores the material compound details. The rest of the fields consist mainly of the paper details.

\subsection{Extracting Triplets}
Every record in the battery database has numerous fields. However, for our task, we confine ourselves to just four of these, namely, \textit{Property-Name}, \textit{Entity-Names}, \textit{Value}, and \textit{Unit}. In the database, the field \textit{Entity-Names} generally contains a chemical compound. \textit{Property-Name} can take up any of the five permissible values `Voltage', `Coulombic Efficiency', `Conductivity', `Capacity', and `Energy'. The \textit{Value} and \textit{Unit} parts contain the value of the \textit{Property} found for the \textit{Name} and its unit, respectively. We generate one triplet from each json entity of the battery database. Traditionally, a triplet consists of two entities and a relation between them. Following the same, \textit{Name} of the json entity becomes the first entity in our triplet. 
\textit{Value} and \textit{Unit} are combined to form the second entity, and \textit{Property} becomes the relation connecting the two entities. The database originally consists of nearly 292,000 entities. However, when we generate triplets, we find that many of them are duplicates, so we perform de-duplication, and the resultant set is what we refer to as the ``set of triplets" from now on.

\subsection{Collecting Articles}
Since the set of triplets is based on sub-domain of battery materials, we require material science corpus to map these triplets to their probable source text. 
Each data record also contains a ``DOI" field that denotes the article that is used to generate the entry. We extract the DOIs from all records in the battery database. Each DOI in this list thus contributes at least one record in the battery database. The list consists of more than 42,000 DOIs. We find most of them on Elsevier\footnote{\url{https://www.elsevier.com/solutions/sciencedirect/librarian-resource-center/api}}, Arxiv\footnote{\url{https://info.arxiv.org/help/api/index.html}} and Springer\footnote{\url{https://www.springeropen.com/get-published/indexing-archiving-and-access-to-data/api}}. We use publicly available APIs of different publishing houses to get the most of the articles. We crawl very few articles which are publicly available and free. Thus, we end up with a corpus of $\sim$40,000 articles in pdf form\footnote{We shall release only sentences with triplets, not the PDF articles.}.

\begin{figure*}[t]
    \centering
    \includegraphics[width=15cm]{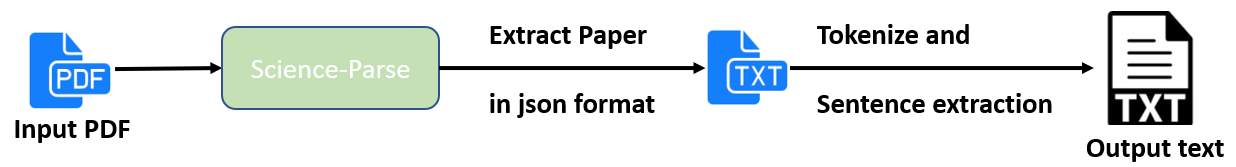}
    \caption{PDF Parsing and Text Extraction.}
    \label{fig:pdf-to-text}
\end{figure*}

\subsection{PDF Parsing and Pre-processing}

To extract the contents of the PDF file, we use `Science-Parse' \footnote{\url{https://github.com/allenai/science-parse}} which parses scientific papers (in PDF form) and retrieves the information in a structured form. It extracts the Title, Authors, Abstract, Sections, Bibliography, Mentions,
etc., in a json format. Moreover, `Science-Parse' is an open-source, accessible, fast tool that extracts text to make the model machine readable. It helps to convert each document in our corpus from pdf to json format. 
After extracting the sentences from json documents, we tokenize using the python NLTK library. The detailed pdf parsing and sentence (text format) extraction steps are shown in Fig. \ref{fig:pdf-to-text}.


\subsection{Using Distant Supervision} 

After collection of triplets and the contents from the PDF files, the mapping is done between a triplet and the corresponding sentence in the PDF file. We use a distant supervision approach to do this. The idea of distant supervision is that if both the entities of a triplet are present in a sentence, we assume that the sentence represents some relation between the entities. In our case, if a sentence contains both entities of a triplet and the relation, it is assumed that the sentence reinforces the same relation as the one present in the triplet. However, this assumption makes the results of distant supervision noisy and often requires a human to judge the output's quality manually.

Distant Supervision helps identify which triplet is a positive example for a given sentence. The first entity of our triplet is a chemical compound, while the second entity is a combination of a value and a unit used to express the relation specified in the triplet.

We use a very strict version of distant supervision that requires both entity parts (first and second entities) and the relation part of the triplet to be present in the sentence. 
Since the second entity consists of scientific units like Amperes and Volts, articles could express the same unit in different notations or forms. It is essential to also look for alternate representations of these units. For example, if the second part of the entity contains `ampere', then we must look for its alternate representations like `A', `amps', `amperes', `ampere', `amp', etc. in the sentence. The process is thus modified to check for alternate representation for all units appearing in the second entity across all triplets. 

Since the second entity always consists of multiple tokens, it is essential to monitor the sequence in which the individual tokens constitute the triplet's second entity. For example, consider the second entity part of a triplet as `150.7 mAh/g', then it could be expressed as `mA h g(-1)' or  `g(-1) mA h' or `mA(1) g(-1) h(1)'. We do not strictly match the entire sequence but check for all possible combinations instead. 

After making the above changes, we successfully generate a set of $\sim$11,000 sentence-triplet pairs over a corpus consisting of $\sim$3,812 research papers. 
These $\sim$11,000 pairs were generated over a set $\sim$6,000 unique sentences. 
This is the `{\sl distantly supervised dataset}'. A sample entry in the dataset is shown below:
From the sentence - 
``Nevertheless, the pure LiCoO2 showed a higher working voltage (3.96 V) than the mixture containing LiNi0.8Co0.17Al0.03O2 and LiCoO2'' - the triplet is: [`LiCoO2', `Voltage', `3.96 V'].

\subsection{Ground Truth Dataset Annotation}

To verify the correctness of the distant supervision based dataset generation and our joint relation-entity extraction model, 
we randomly select 114 papers from the set of 3,812 papers 
and provide them (using tool \cite{mullick2021rte}) to two different material science knowledge experts along with a good working proficiency in English to manually identify the triplets with the five relations (Conductivity, Coulombic Efficiency, Capacity, Voltage, and Energy). The total number of sentences annotated by the experts for these articles is 1,255. We select three annotators after several discussions and conditions of fulfilling many criterias like annotators should have domain knowledge expertise along with a good working proficiency in English. Annotators are experienced in similar tasks earlier also (with earlier publications and datasets are public) so the annotators are reliable in these annotation tasks. Initial labeling is done by two annotators and any annotation discrepancy is checked and resolved by the third annotator after discussing with two others. So, the final judgment is based on three annotators’ discussion. 
One common conflict arises regarding the inclusion of the material name. For example, in the following sentence -``Tin phosphide ( Sn4P3 ) has been a promising anode material for SIBs owing to its theoretical capacity of 1132 mA h g 1 and high electrical conductivity of 30.7 S cm 1 " - One annotation contains "Tin phosphide ( Sn4P3 )" and another contains only "Sn4P3". Finally, "Tin Phosphide" is not included.
 The inter-annotator agreement Cohen $\kappa$ between two annotators is 0.82 including these types of conflicts - which is almost perfect $\kappa$ score as described in \citet{landis1977measurement}. The statistics of the five relations are shown below, which exhibit a count of different relations among entity pairs. 
\begin{itemize}
    \itemsep0em
    \item Conductivity: 122
    \item Coulombic Efficiency: 553
    \item Capacity: 378
    \item Voltage: 637
    \item Energy: 103
\end{itemize}

%% file: 5Approach.tex
\begin{figure*}[htp]
    \centering
    \includegraphics[width=0.9\linewidth]{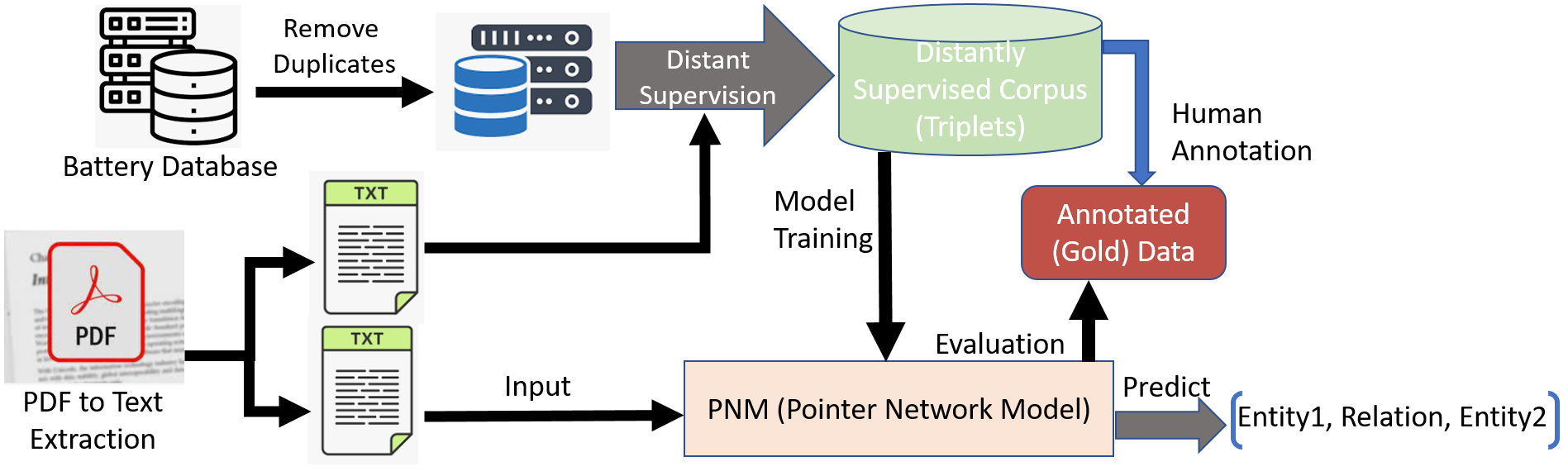}
    \caption{Overall Framework}
    \label{fig:overall-framework}
\end{figure*}

\section{Problem Definition and Proposed Solution}
\label{sec:problem}

To formally define the problem setting, our goal is to predict the relation between two entities in a material science article and extract these in the form of a triplet - $(entity1, relation, entity2)$. As discussed earlier, we first create a corpus with the sentences and corresponding triplets and annotate a part of the dataset. Then, we train a relation extraction model using this corpus, which can predict triplets given a new research article in this domain. Our overall system framework is shown in Fig \ref{fig:overall-framework} where at the first step, we use distant supervision to create a corpus of sentences with triplets and use a subset of this dataset for gold standard annotation. We then train the pointer network model (PNM) and predict the entities and corresponding relations from a given sentence. Annotated data is used to evaluate the algorithm against gold standard data. We describe our model in the next section.

\begin{figure*}[htp]
    \centering
    \begin{adjustbox}{width=0.9\linewidth}
    \includegraphics[width=\linewidth]{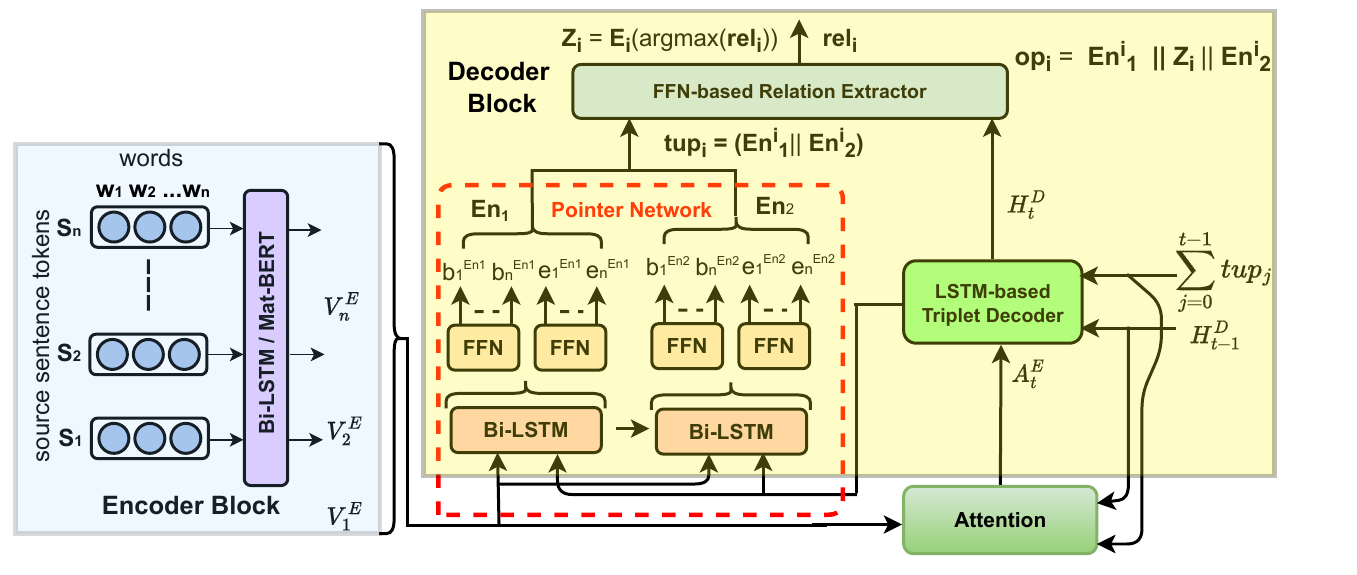}
    \end{adjustbox}
    \caption{Proposed Encoder-Decoder Based Pointer Network Model}
    \label{fig:pointer-network-model}
\end{figure*}

Earlier works can not handle overlapping entity extraction and mostly focus on single relation retrieval at a time, so they would need separate sequence label models for each relation, increasing the time complexity of the model. To overcome these setbacks, we propose using pointer network-based encoder-decoder model, which has the following advantages: it is a joint model for entity extraction and relation classification. It can detect any number of triplets in a sentence, even if there is an overlap between the entities. Our model takes a sentence as input and returns a set of triplets present in the sentence as the output. One triplet consists of two entities and the corresponding relationship between these two. The triplet form is $(entity1 | relation | entity2)$. `|' is the separator token for each component of the triplet. The pointer network based framework is shown in Fig \ref{fig:pointer-network-model} where the input sentence $S_{i}$ = $\{w_{1}, w_{2}, ..., w_{n}\}$ contains $n$ words. The model aims to extract a set of entity-relation triplets as output,
\begin{equation}
	Op = \{op_{i} | op_{i} = [(b_{i}^{p_{1}}, e_{i}^{p_{1}}), Z_{i}, (b_{i}^{p_{2}}, e_{i}^{p_{2}})]\}_{i=1}^{|T|}
\end{equation}
where $op_{i}$ denotes the $i^{th}$ triplet and $|Op|$ denotes the size of the triplet set. 
$b_{i}^{p_{1}}$ and $b_{i}^{p_{2}}$ represents the beginning position of entity 1 and entity 2 respectively for the $i^{th}$ triplet. Similarly, $e_{i}^{p_{1}}$ and $e_{i}^{p_{2}}$ denotes the end position of entity 1 and entity 2 for the $i^{th}$ triplet. So ($b_{i}^{p_{1}}$ and $e_{i}^{p_{1}}$) marks the entity 1 for the $i^{th}$ triplet. Similarly, ($b_{i}^{p_{2}}$ and $e_{i}^{p_{2}}$) marks the entity 2. $Z_{i}$ represents the possible relation between the above two entities. $p_1$ and $p_2$ denote the two pointer network models.

For the same entities with multiple relations, we create separate triplets. We build an encoder-decoder architecture where sequence generation occurs from an input sentence. Relation tuples can be retrieved with the help of a separator token. 
In the encoder-decoder model, firstly, the sentence encoding vectors are generated. Then, with the help of indicator words for identifying relations in a sentence, the decoder model identifies the start and end of a relation to extract the relation and entities. Similar to \citet{nayak2020effective}, we experiment with different encoding and decoding-based models to extract the triplets. 
For decoding the triplets, we mainly use two approaches, which are (i) word-level decoding and (ii) position-based decoding. The encoder modules for these two approaches are the same. We describe the different components of these models below.

\subsection{Encoder}

One sentence can contain one or multiple relations among different entities in that sentence, but the association of a relation between two specific entities is highly contextual. Hence, we explore LSTM-based (Bi-LSTM) and transformer-based models to capture the hidden contextual representation.

We use seven different embeddings in the pointer network encoder: a) Word2Vec+Bi-LSTM model, b) BERT model, c) RoBERTa model, d) SciBERT e) BatteryBERT, f) MatSciBERT model and g) MatBERT model. Word2Vec embeddings are utilized with the Bi-LSTM layer. Different variants of BERT replace Word2Vec+Bi-LSTM for encoding the sentences and obtaining the representation.

\noindent \textbf{Word2Vec+Bi-LSTM:} Training a word-embedding model is required to obtain the word representation matrix. \textit{skip-gram} approach is needed to build the word embedding model. 
The Skip-gram architecture uses the distributional hypothesis assumption, which states that similar words tend to appear in similar contexts. Given a word, it learns the context of that word with respect to surrounding words. We obtain the word representation vector using the \textit{skip-gram} architecture based \textit{word2vec} model as discussed by \citet{mikolov2013efficient}. With the help of Word2Vec, we extract the word vectors of input sentence tokens and various relation names from relation set $R$, different separator tokens (like `|'), and using all, we generate one common vocabulary ($V$). We also add special tokens like - begin of the token sequence ({\em BOT}), the end of the token sequence ({\em EOT}), and the unknown token ({\em UNK}) to $V$. We use a word embedding layer ($\mathbf{Emb}_w \in \mathbb{R}^{\vert V \vert \times D_{wv}}$ and a character embedding layer $\mathbf{Emb}_c \in \mathbb{R}^{\vert N \vert \times D_{ch}}$, where $D_{wv}$ is the dimension of word vectors, $N$ is the number of alphabet characters of the source sentence, and $D_{ch}$ is the dimension of the character embedding vectors. We concatenate the character embedding-based vectors with pre-trained word vectors to obtain the vector form of the source sentence tokens. A Bi-LSTM layer processes the input embeddings to get the contextual representation.

\noindent \textbf{BERT \cite{devlin2019bert} :} BERT stands for `Bidirectional Encoder Representations from Transformers'. It uses Masked Language Modelling(MLM) as the objective for pre-training. MLM randomly masks 15\% of the input text tokens, and it predicts the masked token using the bi-directional context of the original text. Thus, it pretrains a deep bidirectional transformer. It also jointly pretrains the next sentence prediction task. For pre-training, various corpora like BooksCorpus (800M words) \cite{zhu2015aligning}, English Wikipedia (2,500M words), and Word Benchmark \cite{chelba2013one} are used.
We use BERT-base uncased model with ($L=12$, $H=768$, $A=12$, Total Parameters = $110M$).

\noindent \textbf{RoBERTa~\cite{Liu2019RoBERTaAR} :} 
RoBERTa is an updated version of BERT, which overcomes several limitations of the BERT model by tuning hyper-parameters with several updates, using bigger batch size, longer training time with more data, and using different masking patterns dynamically.
For pretraining, a corpora containing five English datasets with various sizes are used, which are - BooksCorpus (800M words) with Wikipedia having 16GB size \cite{zhu2015aligning}, CC-News by Sebastian Nagel in 2016\footnote{http://web.archive.org/save/http://commoncrawl.org/2016/10/newsdataset-available.} with 76GB size, Open Web Text \cite{openwebtext2019} having 38GB size and Stories~\cite{trinh2018simple} with 31GB size. 

\noindent \textbf{SciBERT~\cite{beltagy2019scibert} :} It is a pre-trained language model for scientific text based on BERT. They train their model on various scientific domain datasets with different sizes.

\noindent
\textbf{BatteryBERT~\cite{huang2022batterybert} :} BatteryBERT is a pre-trained language model on Battery data. This model is trained on a corpus of battery research papers.

\noindent
\textbf{MatSciBERT~\cite{gupta2022matscibert} :} MatSciBERT is a pre-trained language model pre-trained on a large number of material science publications.

\noindent \textbf{MatBERT~\cite{walker2021impact} :} A large text corpus of a similar domain is required to build and train the model. 
We apply a pre-trained version of MatBERT as our dataset needs to be more extensive for building a pre-train model. It is a material science domain-specific pre-trained model similar to BERT.

The BERT models use pre-trained BERT embeddings. For different BERT models, the pre-training was done on different datasets. For example, MatBERT is more relevant to material science datasets, and tokenization takes care of this. The tokens consist of material names, for example, Li2O, which are different from ordinary words as they consist of small and large letters with numbers in between. In the case of BERT, the tokens are represented as a sequence of characters. 
Let, $S_i$ be the $i^{th}$ sentence containing $w_1$, $w_2$ ... $w_n$ words. After the sentence encoding, the encoder generates a vector ($\mathbf{V}_i^E$) from the $i^{th}$ sentence $S_i$. It is shown in the `Encoder Block' in Fig \ref{fig:pointer-network-model}.

\subsection{Decoder}
We employ two different models for the decoder - a word decoding model and a pointer net decoding model.

\subsubsection{Word Decoding Model (WDM)}
In this, we use an LSTM decoder for word-level decoding. Let $\mathbf{T}$ be a target token sequence that is formed by word embedding vectors of its word tokens - $\mathbf{tup}_0$, $\mathbf{tup}_1$,....,$\mathbf{tup}_m$ where $\mathbf{tup}_i \in \mathbb{R}^{D_{wv}}$ is the embedding vector of the $i$-th token, $m$ is the length of the target sequence, and $D_{wv}$ is the dimension of word vectors. Two special tokens - {\em BOT} (Begin of Token) and {\em EOT} (End of Token) are denoted by $\mathbf{tup}_0$ and $\mathbf{tup}_m$. At a time, the LSTM decoder produces a single token and halts by {\em EOT} token. At $t^{th}$ time step, `$tup_{t}$' denotes current token's embedding vector and ($A_{t}^E$) use source sentence encoding. Utilizing the source sentence encoding ($A_{t}^E$), the previous target word embedding ($tup_{t-1}$), and the previous hidden vector ($H_{t-1}^D$), the LSTM sequence generator produces a new hidden representation of current token ($H_{t}^D \in $ $\mathbb{R}^{D_h}$). $\mathbb{R}^{D_h}$ represent hidden state dimension. We use an attention mechanism to encode the sentences and generate a sentence vector. 

We use gold-label annotated tokens during training, but the decoder may generate out-of-vocabulary tokens at inference time. To handle this,  a masking technique is introduced along with a special token - {\em UNK}. All words of vocabulary can be masked except - the current tokens of the input sentence, the relation tokens, other separator tokens (`;', `$\vert$'), {\em UNK}, and {\em EOT} tokens. When the decoder generates an {\em UNK} token, it is replaced with the corresponding source word with the highest attention score. 
Once the decoding process is over, in the inference stage, all tuples are extracted after removing the duplicates. We consider this model as the {\em Word Decoding Model} (WDM).

\subsubsection{Pointer Network-based Decoding}

We employ a Pointer Network-based approach to detect two entities and predict the relation between them. First, we detect the entities using their begin and end positions and remove special tokens (as discussed in the previous Wordnet Decoding Model) and relation names from the vocabulary. In the decoder part of the pointer network architecture, we create a relation matrix, $\mathbf{E}_r \in \mathbb{R}^{\vert R \vert \times D_{rel}}$. $R$ is the set of relations and $D_{rel}$ is the dimension of relation vectors. Similar to the earlier WDM model, a special relation token {\em EOT} is present in the relation set $R$. It denotes the end of the token sequence. Relation tuples are represented as a sequence $T=Z_0, Z_1,....,Z_m$, where $Z_t$ is a tuple consisting of four position information in the input sentence marking the beginning and ending indexes of the two entities and the relation between them ($rel_t$). $Z_0$ is a dummy tuple for starting, and $Z_m$ (for m sequence) denotes the end of the sequence by {\em EOT} token. The decoder consists of an LSTM-based triplet sequence generator that produces the intermediate hidden representation of the current tuple ($H_{t}^D$) at timestamp t. Two pointer networks, each consisting of Bi-LSTM and FFN (Feed Forward Networks), are utilized to retrieve the two entities. Thereafter, a feed forward relation extractor operates on two entities and explores their feature inter-dependencies to predict the relation that exists between them. The overall architecture of the pointer network-based decoder is shown in Fig \ref{fig:pointer-network-model}. We discuss different blocks below.

\textbf{Attention Modeling:} We use the attention algorithm proposed by \citet{bahdanau2014neural} which takes previous tuple ($tup_{t-1}$) and hidden vector ($H_{t-1}^D$) as input at timestamp $t$ to produce the attention weighted context vector ($A_{t}^E$) for the current input sentence. 

\textbf{Triplet Sequence Generator:} The triplet sequence generator structure is based on an LSTM layer with hidden dimension $D_h$ to produce a sequence of tuples (two entities) that later form the triplets. At timestamp $t$, it takes the source sentence encoding using attention layer ($A_{t}^E$), the previous tuple description using pointer networks ($\mathbf{tup}_{t-1}=\sum_{j=0}^{t-1}\mathbf{tup}_{j}$) and hidden vectors ($H_{t-1}^D$) as input to generate the hidden representation of current token ($H_{t}^D \in $ $\mathbb{R}^{D_h}$). ($\overrightarrow{0}$) to denote the dummy tuple $tup_0$. The LSTM outcome is as follows:
\begin{align*}
&\mathbf{tup}_{t} = \sum_{j=0}^{t-1} \mathbf{tup}_j, \quad
\mathbf{H}_t^D = \mathrm{LSTM}(\mathbf{A}_t^E \Vert \mathbf{tup}_{t-1}, \mathbf{H}_{t-1}^D)
\end{align*}

\textbf{Pointer Network:} 
A Bi-LSTM layer with hidden dimension $\mathbf{D}_{BH}$, followed by two FFN (Feed Forward Networks), constitutes a pointer network. Here we use two-pointer networks for extracting the two entities. We concatenate $\mathbf{H}_t^D$ and $\mathbf{V}_i^E$ (obtained from the encoding layer) to provide the input of a Bi-LSTM model (forward and backward LSTM), which provides a hidden representation ($H_{i}^m \in $ $\mathbb{R}^{2D_{BH}}$) to be fed to FFN models. Two FFNs with softmax provide scores between 0 and 1, the begin ($b$) and end ($e$) index of one entity. 
\begin{align*}
&\hat{b}_i^1 = \mathbf{W}_b^1 \mathbf{h}_i^m + {bias}_b^1,
\quad \hat{e}_i^1 = \mathbf{W}_e^1 \mathbf{h}_i^m + {bias}_e^1\\
&\mathbf{b}^1 = \mathrm{softmax}(\hat{\mathbf{b}}^1),
\quad \mathbf{e}^1 = \mathrm{softmax}(\hat{\mathbf{e}}^1)
\end{align*}
\noindent where $\mathbf{W}_b^1 \in \mathbb{R}^{1 \times 2D_{BH}}$, $\mathbf{W}_e^1 \in \mathbb{R}^{1 \times 2D_{BH}}$ are the weight parameters of FFN. ${bias}_b^1$ and ${bias}_e^1$ are the bias parameters of the feed-forward layers (FFN). ${b}_i^1$ and ${e}_i^1$ are normalized probabilities of the $i^{th}$ source word. ${b}_i^1$ and ${e}_i^1$ denotes the begin and end token of the first entity of the $i^{th}$ predicted tuple, respectively.

Then, the second pointer network model extracts the second entity. After concatenating the first Bi-LSTM output vector ($\mathbf{H}_i^m$) with triplet decoder output ($\mathbf{H}_t^D$) and sentence encoding ($\mathbf{V}_i^E$), we feed them to the second pointer network to obtain the position of the begin and end tokens of the second entity. Together, these two pointer networks produce the feature vectors $tup_{t}$ containing entity1 ($En_1^t$) and entity2 ($En_2^t$).

\textbf{Relation Extraction:}
We pass $tup_{t}$ and $H_t^D$ to the FFN-based Relation Extraction model, a classifier with softmax function to produce normalized probability, $rel_{t}$ at time step $t$. The existence of a relation in a sentence is generally not dependent on the occurrence of a specific word but rather on a set of clues in entities that the model must learn. The classifier identifies the interactions between two entities and predicts the relation. The final relation is obtained using argmax on $rel_{t}$ and relation extractor ($E_r$). 
\begin{equation}
    Z_{i} = E_r (argmax(rel_t))
\end{equation}
and the final triplet output is represented as,
\begin{equation}
    op_t = (En_1^t || Z_i || En_2^t)
\end{equation}

Once the dataset is created, for all the files containing the train, test, and validation sentences, two files, .sent and .pointer, are generated. The .pointer file stores the position of the material name, value, and relation. For a sentence, this file stores the beginning and ending indices of the material name, the beginning and ending locations of the value, and then in the middle, the corresponding relation name. Table \ref{fig:pnm-sample-op} shows an example of pointer-network-based decoding of the source sentence (from the manuscript of
\citet{yang2016cu0}).

\begin{table*}[!thb]
\centering
\caption{ Relation tuple representation of pointer-network model }
\label{fig:pnm-sample-op}
\begin{adjustbox}{width=0.9\linewidth}
\begin{tabular}{|c|c|}
\hline
Source Sentence & However, at the same C-rate, the Cu0.02Ti0.94Nb2.04O7 \\ &sample exhibits a larger first-cycle Coulombic efficiency\\& ( 91.0 \% ) than that of the TiNb2O7 sample ( 81.6 \% ) probably\\& due to the smaller particle size and larger ( electronic\\& and ionic ) conductivity of Cu0.02Ti0.94Nb2.04O7 [ 6,38 ].
\\ \hline
Pointer-network-based decoding & 
<8 8 17 18 Coulombic\_Efficiency>
| <24 24 27 28  Coulombic\_Efficiency>
\\ \hline
\end{tabular}
\end{adjustbox}
\end{table*}

\begin{table*}[b]
\centering
\begin{adjustbox}{width=0.85\linewidth}
\begin{tabular}{|c|c|c|c|c|c|c|}
\hline
 & \multicolumn{3}{c|}{\textbf{Distantly Supervised Dataset}} & \multicolumn{3}{c|}{\textbf{Annotated Ground Truth Dataset}} \\\hline
 & \textbf{Pr, $sd$} & \textbf{Re, $sd$} & \textbf{F1, $sd$} & \textbf{Pr, $sd$} & \textbf{Re, $sd$} & \textbf{F1, $sd$}\\ \hline
WDM & 0.808, 0.009 &
0.615, 0.005 &
0.698, 0.016 &
0.469, 0.013 &
0.410, 0.006 &
0.439, 0.008\\ \hline
DyGIE++ & 0.865, 0.008 & 0.909, 0.007 & 0.886, 0.007 & 0.869, 0.006 & 0.910, 0.005 & 0.889, 0.006\\ \hline
MatKG & 0.892, 0.007 & 0.903, 0.008 & 0.897, 0.007 & 0.898, 0.009 & 0.906, 0.007 & 0.901, 0.008\\ \hline
PNM (Word2Vec) & 0.884, 0.028 &
0.848, 0.013 &
0.866, 0.005 &
0.856, 0.007 &
0.680, 0.030  &
0.758, 0.009\\\hline
PNM (RoBERTa) & 0.821, 0.022 & 0.782, 0.007 &
0.801, 0.003 & 0.837, 0.007 & 0.765, 0.018 & 0.798, 0.007\\ \hline
PNM (SciBERT) &  0.905, 0.010 &
0.874, 0.006 &
0.890, 0.019 &
0.894, 0.003 &
0.880, 0.015 &
0.885, 0.009\\ \hline
PNM (BERT) & 0.921, 0.024 &
0.883, 0.003 &
0.902, 0.014 &
0.926, 0.003 &
0.860, 0.015 &
0.908, 0.006 \\ \hline
PNM (BatteryBERT) & 0.905, 0.011 &
0.878, 0.006 &
0.891, 0.009 &
0.886, 0.006 &
0.894, 0.007 &
0.890, 0.009 \\ \hline
PNM (MatSciBERT) & 0.916, 0.009 &
0.881, 0.010 &
0.898, 0.008 &
0.904, 0.007 &
0.873, 0.012 &
0.889, 0.008 \\ \hline
PNM (MatBERT) & 0.920, 0.008 &
0.907, 0.004 & \textbf{0.913}, 0.002 & 0.919, 0.005 & 0.911, 0.004 & \textbf{0.915}, 0.006\\ \hline
\end{tabular}
\end{adjustbox}
    \caption{Weighted Average scores of Precision (Pr), Recall (Re), F1-Score (F1) and their respective standard deviations ($sd$) of different models on distantly supervised dataset and annotated ground truth dataset}
  \label{tab:all-macro-on-all-data}
\end{table*}

The output shows the generation of two triplets of the form entity1-relation-entity2 from the sentence. Entity1 is the material name. For both the triplets, the start and end position of the entity1 names (Cu0.02Ti0.94Nb2.04O7 and TiNb2O7) can be confirmed as 8 and 24, respectively. Similarly, the start and end positions of entity2 (91.0 \% and 81.6 \%) are 17 and 18 for the first triplet and 27 and 28 for the second triplet, respectively. Finally, the relation name, coulombic efficiency, is stored.

After running the model, the output should be as follows for two different entity-relation triplets: 
\begin{itemize}
  \item Prediction 1: Cu0.02Ti0.94Nb2.04O7, Coulombic Efficiency, 91.0\% 
  \item Prediction 2: TiNb2O7, Coulombic Efficiency, 81.6 \%
\end{itemize}

\subsection{Other Baselines:}
\noindent \textbf{A) DyGIE++} (\citet{wadden2019entity}): 
It is a multi-task framework designed for: named entity recognition, relation extraction and event extraction. It is implemented by constructing text span representations and scoring the same to capture local and global context. For an entity-relation framework, a span graph is generated from the entity-relation triples from the sentence.

\noindent \textbf{B) MatKG} (\citet{venugopal2022matkg}):
MatKG is a large knowledge graph of material science concepts. It consists of more than 2 million unique relationship triplets (entity-relation-entity) which are derived from 80,000 entities.

We compare performances of DyGIE++ and MatKG with different models on our datasets and evaluations are in Table \ref{tab:all-macro-on-all-data}.

\textbf{C) ChemDataExtractor~\cite{huang2020database}:} We also show comparison with ChemDataExtractor in Table \ref{tab:chemData-compare}.

%% file: 6Experiment.tex
\begin{table*}[!htb]
\centering
\begin{adjustbox}{width=0.85\linewidth}
\begin{tabular}{|c|c|c|c|c|c|c|}
\hline
 & \multicolumn{3}{c|}{\textbf{Distantly supervised dataset}} & \multicolumn{3}{c|}{\textbf{Annotated ground truth dataset}} \\\hline
& \textbf{Pr, $sd$} & \textbf{Re, $sd$} & \textbf{F1, $sd$} &  \textbf{Pr, $sd$} & \textbf{Re, $sd$} & \textbf{F1, $sd$} \\ \hline

\textit{\textit{Voltage}} & 0.918,0.006 &
0.893,0.005 &
0.906,0.005 &
0.922,0.003 &
0.898,0.013 &
0.912,0.008 \\ \hline
\textit{\textit{Capacity}} & 0.914,0.029 &
0.901,0.004 &
0.907,0.015 &
0.885,0.008 &
0.901,0.002 &
0.893,0.012 \\ \hline
\textit{\textit{Conductivity}} & 0.875,0.007 &
0.859,0.009 &
0.866,0.002 &
0.868,0.034 &
0.868,0.007 &
0.868,0.004 \\ \hline
\textit{\textit{Coulombic Efficiency}} & 0.925,0.008 &
0.919,0.003 &
0.921,0.006 &
0.933,0.006 &
0.922,0.003 &
0.927,0.005 \\ \hline
\textit{\textit{Energy}} & 0.905,0.006 &
0.942,0.012 &
0.923,0.009 &
0.911,0.015 &
0.953,0.006 &
0.932,0.008 \\ \hline
\textbf{macro score} & 0.920,0.008 &
0.907,0.004 &
\textbf{0.913},0.002 & 0.919,0.005 &
0.911,0.004 &
\textbf{0.915},0.007\\ \hline
\end{tabular}
\end{adjustbox}
    \caption{Precision (Pr), Recall (Re), F1-Score (F1) and their respective standard deviations ($sd$) of PNM (MatBERT) on distantly supervised dataset and annotated ground truth dataset}
  \label{tab:bert-mat-annnot}

\end{table*}

\section{Experimental Results}
\label{sec:experiment}

To validate the proposed model, we first compare the proposed Pointer Net Decoding Model (PNM) while taking various embeddings as input: Word2Vec, BERT, SciBERT, MatBERT, RoBERTa, MatSciBERT, and BatteryBERT on both the distantly supervised corpus and the ground truth annotated dataset. We then compare the proposed model with the ChemDataExtractor model on a manually annotated subset of the battery dataset~\cite{huang2020database}. After that, we experimented with the MatBERT model while varying the training data size to understand how much training data is required for decent performance. We also perform few-shot experiments to check the approach's usefulness in the presence of minimal data. 

Next, we describe the evaluation metrics used.

\begin{table*}[b]
\centering
\begin{adjustbox}{width=0.85\linewidth}
\begin{tabular}{|c|c|c|c|c|c|c|c|c|c|c|c|c|}
\hline
 & \multicolumn{3}{c|}{\textbf{10\%}} & \multicolumn{3}{c|}{\textbf{30\%}} & \multicolumn{3}{c|}{\textbf{50\%}} & \multicolumn{3}{c|}{\textbf{70\%}}\\\hline
 & \textbf{Pr} & \textbf{Re} & \textbf{F1} &  \textbf{Pr} & \textbf{Re} & \textbf{F1} & \textbf{Pr} & \textbf{Re} & \textbf{F1} &  \textbf{Pr} & \textbf{Re} & \textbf{F1} \\ \hline
\textit{\textit{Voltage}} & 0.874 &	0.777 &	0.823 & 0.913 &	0.855 &	0.883 & 0.904 & 0.869 & 0.886 & 0.904 & 0.879 & 0.892 \\ \hline
\textit{\textit{Capacity}} & 0.848 & 0.806 & 0.827 & 0.878 & 0.856 & 0.867 & 0.896 & 0.838 & 0.866 & 0.909 & 0.881 & 0.895\\ \hline
\textit{\textit{Conductivity}} & 0.778 & 0.396 & 0.525 & 0.755 & 0.755 &	0.755 & 0.812 & 0.736 & 0.772 & 0.83 & 0.736 & 0.78 \\ \hline
\textit{\textit{Coulombic Efficiency}} & 0.814 & 0.826 & 0.82 & 0.839 &	0.877 & 0.858 & 0.869 & 0.898 & 0.883 & 0.891 & 0.88 & 0.885\\ \hline
\textit{\textit{Energy}} &  0.704 &	0.581 &	0.637 & 0.948 &	0.849 &	0.896 & 0.975 & 0.919 & 0.946 & 0.953 & 0.942 & 0.947\\ \hline
\textbf{macro score} & 0.857 &	0.781 &	0.817 & 0.897 &	0.863 &	0.879 & 0.902 & 0.874 & 0.888 & 0.907 & 0.887 & 0.897\\ \hline
\end{tabular}
\end{adjustbox}
\caption{Precision (Pr), Recall (Re), F1-Score (F1) for PNM (MatBERT) for 10\%, 30\%, 50\%, and 70\% of usage training samples during training phase for distantly supervised dataset}
\label{tab:train-vary}
\end{table*}

\subsubsection{Evaluation Metrics}
This section describes the specific metrics we use to measure performance. Relation identification can be thought of as a multi-class classification task. 
We apply similar evaluation metrics - Precision (Pr), Recall (Re), and F1-Score (F1) as in \citet{nayak2020effective}. Precision ($Pr$) corresponding to a particular relation can be expressed as 

\begin{equation}
Pr = \frac{\mid tr \cap p \mid}{\mid p \mid}
\end{equation}

where \(tr\) represents the set of triplets that actually belong to the relation, and \(p\) represents the set of triplets that are predicted to correspond to the given relation by our model. It must be noted that the numerator of the above expression would contain a triplet if and only if both entities and the relation are predicted correctly by the model. Recall ($Re$) can be defined by the notation - 
\begin{equation}
Re = \frac{\mid tr \cap p \mid}{\mid tr \mid}
\end{equation}

$F1$-score is defined by the harmonic mean of precision and recall, i.e., 
\begin{equation}
F1 = \frac{2Pr*Re}{Pr + Re}
\end{equation}

A higher $F1$-Score is desirable for a model.

\begin{equation}
sd =\sqrt{\frac{1}{N-1}\sum_{i=1}^{N} (X_{i} -\bar{X})^2} 
\label{eq:sd}
\end{equation}

In our experiments, we calculate the weighted average precision (Pr), recall (Re), and F1-score (F1) along with their respective standard deviations ($sd$) 
where standard deviation values are calculated as per Equation \ref{eq:sd}. 

\subsection{Results}

The results are shown for both the distantly supervised corpus and annotated ground truth dataset. 
We divide both the datasets into train, dev, and test. We randomly sample 70\% samples to train the model. A 10\% sample is used for the model parameter and hyper-parameter tuning. The remaining 20\% dataset is used as a blind test set for model evaluation. 
We perform the experiments five different times with different non-overlapping blind test sets and report the weighted average precision (Pr), recall (Re), and F1-score (F1) along with their respective standard deviations ($sd$) for each of the five different relations.

Table \ref{tab:all-macro-on-all-data} shows weighted average precision, recall, and F1-score and respective standard deviations for Word Decoding Model (WDM), DyGIE++, MatKG and Pointer Net Decoding Model (PNM) using different embeddings on both the distantly supervised dataset and the annotated ground truth dataset. Seven different embeddings are utilized in the pointer network encoder: a) Word2Vec+Bi-LSTM model, b) BERT model, c) RoBERTa model, d) SciBERT e) BatteryBERT, f) MatSciBERT model and g) MatBERT model. Results (weighted average and standard deviation) are shown for five relations - voltage, capacity, conductivity, coulombic efficiency, and energy. Pointer Net Model with MatBERT [PNM (MatBERT)] provides the best macro F1-score, followed by PNM (BERT) model. PNM (MatBERT) produced 0.913 and 0.915 macro F1 scores for the distantly supervised and annotated ground truth datasets. PNM (MatBERT) achieves 0.920 precision and 0.907 recall for distantly supervised data and 0.919 precision and 0.911 recall for annotated ground truth dataset. PNM (MatBERT) also has the least standard deviation for average macro precision, recall, and F1 score. DyGIE++ achieves F1-score of 0.886 and 0.889 for distantly supervised and annotated dataset whereas MatKG provides F1-score of 0.897 and 0.901 for distantly supervised and annotated dataset.
Pointer Network Model (PNM) outperforms Word Decoding Model (WDM) for all embeddings and PNM (MatBERT) produces better outcomes than DyGIE++ and MatKG - it shows the superiority of the PNM model. The distantly supervised dataset also achieves similar performance compared to the annotated ground truth dataset using the PNM model - which shows the utility of a distant supervised approach to generate a dataset with less annotation cost.

\begin{table*}[!htb]
\centering
\begin{adjustbox}{width=0.85\linewidth}
\begin{tabular}{|c|c|c|c|c|c|c|c|c|c|c|c|c|}
\hline
 & \multicolumn{3}{c|}{\textbf{10\%}} & \multicolumn{3}{c|}{\textbf{30\%}} & \multicolumn{3}{c|}{\textbf{50\%}} & \multicolumn{3}{c|}{\textbf{70\%}}\\\hline
 & \textbf{Pr} & \textbf{Re} & \textbf{F1} &  \textbf{Pr} & \textbf{Re} & \textbf{F1} & \textbf{Pr} & \textbf{Re} & \textbf{F1} &  \textbf{Pr} & \textbf{Re} & \textbf{F1} \\ \hline
\textit{\textit{Voltage}} & 0.791 &	0.682 &	0.729 & 0.838 &	0.802 &	0.820 & 0.867 & 0.831 & 0.848 & 0.852 & 0.840 & 0.846 \\ \hline
\textit{\textit{Capacity}} & 0.755 & 0.781 & 0.767 & 0.803 & 0.762 & 0.782 & 0.837 & 0.853 & 0.844 & 0.841 & 0.881 & 0.861\\ \hline
\textit{\textit{Conductivity}} & 0.654 & 0.289 & 0.416 & 0.708 & 0.591 & 0.643 & 0.785 & 0.703 & 0.742 & 0.802 & 0.736 & 0.770 \\ \hline
\textit{\textit{Coulombic Efficiency}} & 0.765 & 0.808 & 0.786 & 0.758 & 0.831 & 0.792 & 0.814 & 0.833 & 0.823 & 0.845 & 0.829 & 0.837\\ \hline
\textit{\textit{Energy}} &  0.719 &	0.525 &	0.598 & 0.749 &	0.601 &	0.664 & 0.895 & 0.654 & 0.735 & 0.827 & 0.712 & 0.764\\ \hline
\textbf{macro score} & 0.737 & 0.617 &	0.659 & 0.770 &	0.717 &	0.740 & 0.839 & 0.775 & 0.798 & 0.833 & 0.799 & 0.816\\ \hline
\end{tabular}
\end{adjustbox}
\caption{Precision (Pr), Recall (Re), F1-Score (F1) for PNM (MatBERT) for 10\%, 30\%, 50\%, and 70\% of usage training samples during training phase for ground truth annotated dataset}
\label{tab:train-vary-annot}
\end{table*}

The detailed relation-specific precision, recall, F1, along with their standard deviations ($sd$) for the best model [PNM (MatBERT)] is shown in Table \ref{tab:bert-mat-annnot}. The results suggest that the PNM (MatBERT) model can extract entities and predict relations with good precision, recall, and macro F1-score. 
The same for PNM (BERT) model is shown in the Supplementary Information.

\begin{table*}[b]
\centering
\begin{adjustbox}{width=0.85\linewidth}
\begin{tabular}{|c|c|c|c|c|c|c|c|c|c|c|c|c|}
\hline
& \multicolumn{6}{|c|}{\textbf{Distantly supervised dataset}} & \multicolumn{6}{|c|}{\textbf{Ground truth annotated dataset}} \\ \hline
& \multicolumn{3}{c|}{\textbf{5 shot}} & \multicolumn{3}{c|}{\textbf{10 shot}} & \multicolumn{3}{c|}{\textbf{5 shot}} & \multicolumn{3}{c|}{\textbf{10 shot}}\\\hline
& \textbf{Pr} & \textbf{Re} & \textbf{F1} & \textbf{Pr} & \textbf{Re} & \textbf{F1} & \textbf{Pr} & \textbf{Re} & \textbf{F1}
& \textbf{Pr} & \textbf{Re} & \textbf{F1}\\ \hline
\textit{\textit{Voltage}} & 0.668 &	0.464 &	0.548 & 0.78 & 0.637 & 0.701 & 0.631 & 0.488 & 0.547 & 0.815 & 0.624 & 0.712 \\\hline
\textit{\textit{Capacity}} & 0.702 & 0.694 & 0.698 & 0.848 & 0.82 & 0.834 & 0.713 & 0.677 & 0.695 & 0.823 & 0.805 & 0.814\\ \hline
\textit{\textit{Conductivity}} & 0.212 & 0.679 & 0.323 & 0.784 & 0.755 & 0.769 & 0.307 & 0.524 & 0.359 & 0.698 & 0.759 & 0.728\\ \hline
\textit{\textit{Coulombic Efficiency}} & 0.69 & 0.655 & 0.672 & 0.79 & 0.736 & 0.762 & 0.614 & 0.597 & 0.605 & 0.824 & 0.745 & 0.784\\ \hline
\textit{\textit{Energy}} & 0.378 &	0.686 &	0.488 & 0.701 & 0.791 & 0.743 & 0.341 & 0.609 & 0.426 & 0.682 & 0.766 & 0.719\\ \hline
\textbf{macro score} & 0.625 & 0.547 & 0.583 & 0.793 & 0.695 & 0.741 & 0.58 & 0.562 & 0.571 & 0.802 & 0.687 & 0.738\\ \hline
\end{tabular}
\end{adjustbox}
\caption{Precision (Pr), Recall (Re), F1-Score (F1) for PNM (MatBert) for 5-shot and 10-shot for both distantly supervised dataset and ground truth annotated dataset}
  \label{tab:k-shots}
\end{table*}

The different parameters and hyper-parameters of the PNM (MatBERT) which perform the best are - "learning\_rate" = "0.001", "optimizer" = "Adam", "dropout" = "0.5", "hidden\_dim" = "300", num\_epochs = "50", "batch\_size" = "32".

\begin{table}[]
\centering
\captionsetup{justification=centering}
\begin{adjustbox}{width=0.95\linewidth}
\begin{tabular}{|c|c|c|c|c|c|c|}
\hline
& \multicolumn{3}{c|}{\textbf{ChemDataExtractor}} & \multicolumn{3}{c|}{\textbf{PNM (MatBERT)}} \\\hline
& \textbf{Pr} & \textbf{Re} & \textbf{F1} & \textbf{Pr} & \textbf{Re} & \textbf{F1}\\ \hline
\textit{\textit{Voltage}} & 0.829 & 0.687 & 0.751 & 0.79 & 0.81 & 0.8\\\hline
\textit{\textit{Capacity}} & 0.793 & 0.574 & 0.666 & 0.826 & 0.771 & 0.797\\ \hline
\textit{\textit{Conductivity}} & 0.759 & 0.526 & 0.621 & 0.684 & 0.754 & 0.717\\ \hline
\textit{\textit{Coulombic Efficiency}} & 0.793 & 0.68 &	0.732 & 0.812 & 0.725 & 0.766\\ \hline
\textit{\textit{Energy}} & 0.755 & 0.729 &	0.742 & 0.721 & 0.808 & 0.762\\ \hline
\textbf{macro score} & 0.803 & 0.646 & 0.716 & 0.768 & 0.774 & \textbf{0.771} \\ \hline
\end{tabular}
\end{adjustbox}
\caption{Precision (Pr), Recall (Re), F1-Score (F1) of PNM (MatBert) \\and ChemDataExtractor model on Battery Database}
  \label{tab:chemData-compare}
\end{table}

We show experimental outcomes on distantly supervised corpus and ground truth annotation data earlier. To compare with other similar approaches, we evaluate the PNM (MatBERT) model and ChemDataExtractor on battery material database~\cite{huang2020database}. The Battery Database used for this comparison consists of 51 papers\footnote{as mentioned in the paper \citet{huang2020database} - dataset \href{https://static-content.springer.com/esm/art\%3A10.1038\%2Fs41597-020-00602-2/MediaObjects/41597_2020_602_MOESM1_ESM.xlsx}{link}}. The class distribution is as follows: 795 capacity relations, 51 conductivity relations, 69 coulombic efficiency relations, 489 voltage relations, and 75 energy relations, which forms a total of 1479 entity-relation triplets. The model outputs are compared with the ground truth outputs from the papers. Table \ref{tab:chemData-compare} shows the 
comparison of ChemDataExtractor~\cite{huang2020database} with PNM (MatBERT), where the latter outperforms the former in overall F1-Score. ChemDataExtractor has better precision but poor recall, downgrading the overall F1 score.

\subsubsection{Varying Training Data Size}

To understand the effect of training data size on the performance, we vary the distantly supervised training data size for the PNM (MatBert) model - taking 10\%, 30\%, 50\%, and 70\% of the initial training data, and train the PNM (MatBERT) model with same parameter and hyper-parameter settings. The results for different training sizes on the distantly supervised dataset and ground truth annotated dataset are shown in Tables \ref{tab:train-vary} and \ref{tab:train-vary-annot} respectively. The annotated ground truth dataset has lower precision, recall, and F1-score values as the training data is very small. We see that with increasing the dataset size, the performance of PNM (MatBERT) improves in terms of precision (Pr), recall (Re), and F1-score (F1). Even at 30\% of the dataset, the model is able to achieve a very good performance - macro F1-score of 0.879 compared to 0.915 for the entire distantly supervised training dataset (0.897 precision and 0.863 recall). For 10\% of the dataset, the model also accomplishes decent precision, recall, and macro F1-score. It shows the effectiveness of the Pointer Network model [PNM (MatBERT)].

\begin{table*}[b]
\vspace{-2mm}
\centering
\caption{The ground truth and prediction output are shown for the paper by \citet{yu2016atomic}}
\label{tab:op-single-paper}
\begin{adjustbox}{width=\linewidth}
\begin{tabular}{|c|c|c|c|}
\hline
\textbf{Relation}   & \textbf{Sentence}  & \textbf{Ground truth} & \textbf{Prediction Output}
 \\ \hline
Capacity&
With a high S loading of 59 wt\%, the NG/S electrode with 20TiO2 cycle coating delivered &
NG/S, 1070 mAh g-1&
NG/S, 1070 mAh g-1\\
& a high specific capacity of 1070 mAh g 1 in the initial cycle at 1C and it remained at 918 mAhg1 after & & \\
& 500 cycles , demonstrating its excellent electrochemical performance as a cathode material for Li–S batteries & & \\ \hline
 Capacity&
After more than 70 cycles at various current densities , the NG/S–20TiO2 &
NG/S–20TiO2,&
NG/S–20TiO2,\\
& electrode still has a reversible capacity of 1237 mAh g 1 when further cycled at 0.1C. & , 1237 mAh g-1& 1237 mAh g-1
\\ \hline
Capacity&
After 500 charge–discharge cycles , the NG / S–20 electrode maintains a high discharge capacity &
NG/S–20TiO2,&
NG/S–20,\\
& of 918.3 mAh g 1, with a high capacity retention of 86\%. &  918.3 mAh g-1&  918.3 mAh g-1\\ \hline
Energy&
Assuming the complete reaction of metallic Li and elemental S to form Li2S , a high theoretical &
Li2S ||| 2600 Wh kg-1&
Li2S ||| 2600 Wh kg-1 \\
& energy density ( over 2600 Wh kg 1 ) could be delivered , 3–5 times higher than those reported for & &\\
& conventional lithium insertion hosts. & &
\\ \hline
Energy&
The results revealed that the binding energy of the sulfur containing species to the anatase-TiO2 &
anatase-TiO2(101), 2.30 eV&
anatase-TiO2(101), 2.18 eV \\
& (101) surface was about 2.30 eV , a little bit lower for the rutile-TiO2(110) surface (2.18 eV).& rutile-TiO2(110), 2.18 eV& rutile-TiO2(110), 2.18 eV
\\ \hline
 Energy&
 The results reveal that the binding energy of the S containing species to ZnO could &
ZnO, 5.40 eV&
ZnO, 5.40 eV\\
& be as high as 5.40 eV , which was substantially larger than that of TiO2 .& & \\ \hline
Voltage&
The comparatively low voltage profile ( average of around 2.2 V ) of Li–S &
Li–S, 2.2 V&
Li–S, 2.2 V\\
& batteries can be well compensated by their high energy density .& &\\ \hline
Conductivity&
For the NG / S–40TiO2 electrode showing slightly inferior performance, the reason could be that &
TiO2, ${10}{-12}$ S $cm{-1}$&
TiO2, ${10}{-12}$ S $cm{-1}$ \\
& TiO2 has intrinsically low Li-ion diffusivity ( ${10}^{-12}$  to ${10}^{-9}$ S $cm{-1}$ ) and electronic & TiO2, ${10}{-9}$ S $cm{-1}$& TiO2, ${10}{-9}$ S $cm{-1}$\\
& conductivity ($10^{-12}$ to ${10}^{-7}$ S $cm{-1}$), which could offset the positive effect. & TiO2, ${10}{-7}$ S $cm{-1}$ & TiO2, ${10}{-7}$ S $cm{-1}$\\ \hline

\end{tabular}
\end{adjustbox}
\end{table*}

\subsubsection{Few Shot Experiments}

Model learning can be very efficient in data-intensive applications but is often hampered when the data set is small. In this scenario, Few-Shot Learning (FSL) method is very useful to tackle this problem. The idea of FSL is to explore the prior knowledge using only a few samples with supervised information and thereafter rapidly generalize to the original problem. We review the utility of PNM (MatBERT) model in few shot settings. For an example, $k$-shot is an FSL settings where only $k$ samples are available for each category to train the model. We investigate FSL in case of $k$-shot settings where $k = 5$ and $10$ and the performance of $5$-shot and $10$-shot approaches for PNM (MatBERT) model on both distantly supervised data and ground truth annotated data are shown in Table \ref{tab:k-shots}. It shows that even with very little amount of data (25 triplets for 5-shot and 50 triplets for 10-shot) the model can also detect the relations. PNM (MatBERT) shows 0.583 macro F1-score (0.625 precision, 0.547 recall) for 5-shot on distantly supervised dataset, 0.571 macro F1-score (0.58 precision, 0.562 recall) for the same on ground truth annotated dataset. For 10-shot approach, PNM (MatBERT) shows 0.741 macro F1-score (0.793 precision and 0.695 recall) on distantly supervised dataset, 0.692 macro F1-score (0.733 precision, 0.65 recall) on ground truth annotated dataset. For 10-shot setting the model can predict relation triplets with decent accuracy, which is quite promising.

%% file: 7sampleop.tex
\section{Data and Software Availability}

We experiment on NVIDIA Tesla K40m GPU with 12GB RAM, 6 Gbps clock cycle, and GDDR5 memory. All methods took less than 4 GPU hours for training. For running the BERT models, we use CUDA version V11.1.105. We use open source software languages like `Python,' and packages like `Pytorch' (pytorch-transformers-1.2.0), `recordclass' (recordclass-0.17.2), etc. for implementation. \footnote{The details of open source softwares in our system are in \url{https://github.com/MatSciRE/Material_Science_Relation_Extraction}}. 
Different Python libraries like - `numpy', `pandas', `pickle' etc., are also utilized. We provide our code, data and curation details of our system in the same GitHub page\footnote{\url{https://github.com/MatSciRE/Material_Science_Relation_Extraction}}.  

\subsection{Demonstration outputs}
We built a sample web application using Python streamlit library deployed on Huggingface Spaces to demonstrate our approach. The demo app can be found in the github webpage. 
Using our approach, the app extracts the relation triples, where a material science paper can be uploaded in PDF format.

\begin{figure}[!htp]
    \centering
    \adjustimage{width=9cm, height=17cm, right}{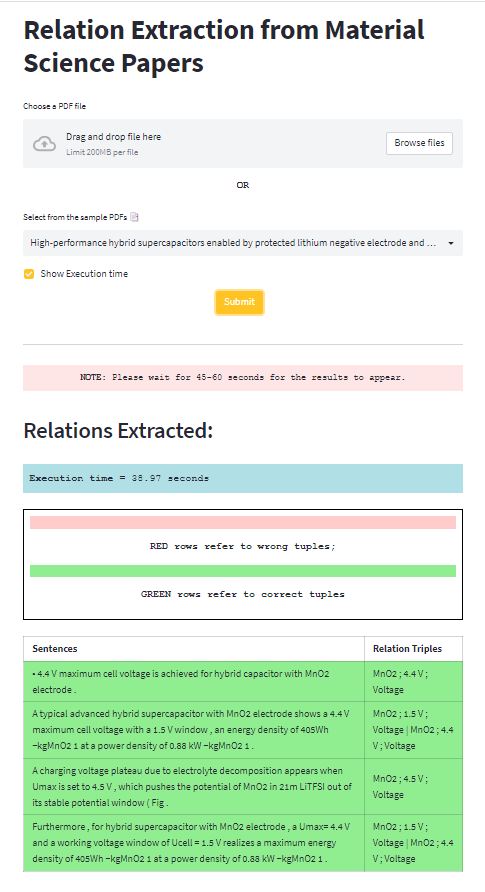}
    \caption{Application demonstrating relation extraction}
    \label{fig:demo-image}
\end{figure}

Figure \ref{fig:demo-image} shows the outputs for the paper "High-performance hybrid supercapacitors enabled by protected lithium negative electrode and 'water-in-salt' electrolyte". The sentences, along with the triplets, are correctly extracted. 

\begin{table*}[!htb]
\centering
\vspace{-2mm}
\caption{Ground truth and prediction output for different relations}
\vspace{-1mm}
\label{tab:op-different-relations}
\captionsetup{justification=centering,margin=0.0mm}
\begin{adjustbox}{width=\linewidth}
\begin{tabular}{|c|c|c|c|}
\hline
\textbf{Relation}   & \textbf{Sentence}  & \textbf{Ground truth} & \textbf{Prediction output}\\ \hline
Capacity&
The NiS–CNT nanocomposite electrode delivers an initial discharge specific capacity of &
NiS–CNT, &
NiS–CNT,\\
& 908.7 mA h g 1  with a coulombic efficiency of 62.5\% in the all-solid-state lithium batteries. &908.7 mA h g 1 & 908.7 mA h g 1
 \\ \hline
 Coulombic&
However, at the same C-rate , the Cu0.02Ti0.94Nb2.04O7 sample exhibits a larger first-cycle &
Cu0.02Ti0.94Nb2.04O7,&
TiNb2O7,
\\
 Efficiency& Coulombic efficiency ( 91.0\% ) than that of the TiNb2O7 sample ( 81.6\% ) probably due to the &  91.0\%&  91.0\%
\\
& smaller particle size and larger (electronic and ionic) conductivity of Cu0.02Ti0.94Nb2.04O7 [ 6,38 ]. & &\\ \hline

 Coulombic&
However, at the same C-rate , the Cu0.02Ti0.94Nb2.04O7 sample exhibits a larger first-cycle &
TiNb2O7, &
TiNb2O7, 
\\
 Efficiency & Coulombic efficiency ( 91.0\% ) than that of the TiNb2O7 sample ( 81.6\% ) probably due to the & 81.6\%& 91.0\% 
\\
& smaller particle size and larger (electronic and ionic) conductivity of Cu0.02Ti0.94Nb2.04O7 [ 6,38 ]. & &\\ \hline

Voltage&
In addition , the discharge capacity of LiNi0.6Co0.2Mn0.2O2 can be &
LiNi0.6Co0.2Mn0.2O2, &
LiNi0.6Co0.2Mn0, \\
&  further increased when the charge cutoff voltage increases to 4.5 V . &4.5V & 4.5V\\ \hline
Conductivity&
Tin phosphide ( Sn4P3 ) has been a promising anode material for SIBs owing to its theoretical &
Sn4P3, 30.7 S cm 1&
Sn4P3, 30.7 S cm 1 \\
& capacity of 1132 mA h g 1 and high electrical conductivity of 30.7 S cm 1 . & &\\ \hline
Energy&
The energy density based on AC and nanowire Na0.35MnO2 &
Na0.35MnO2, 42.6 Wh kg 1&
Na0.35MnO2, 42.6 Wh kg 1 \\
& is 42.6 Wh kg 1 at a power density of 129.8 W kg 1 . & &\\ \hline
\end{tabular}
\end{adjustbox}
\vspace{-2mm}
\end{table*}

\section{Model Output Analysis and Discussions}
\label{sec:sample-op}
We analyze the actual outcomes of the MatSciRE framework, consisting of the PNM (MatBERT) model. For an input sentence from the annotated dataset, we compare the ground truth output with the output generated by the model. The following example from \citet{yang2016cu0} (referenced previously in Table 1) shows the entities and corresponding relations predicted by PNM (MatBERT) from the sentence, compared with the expected (ground truth) output. More examples consisting of other relations are provided in the Supplementary Information.

\textit{However, at the same C-rate, the Cu0.02Ti0.94Nb2.04O7 sample exhibits a larger first-cycle Coulombic efficiency (91.0\%) than that of the TiNb2O7 sample (81.6\%) probably due to the smaller particle size and larger (electronic and ionic) conductivity of Cu0.02Ti0.94Nb2.04O7 [6, 38].}

\textbf{Ground truth:} \newline
<Cu0.02Ti0.94Nb2.04O7, Coulombic efficiency, 91.0\%> 
\newline <TiNb2O7, Coulombic efficiency, 81.6\%>

\textbf{MatBERT(PNM) model output:} \newline <TiNb2O7, Coulombic efficiency, 91.0\%>

It is observed that in relations where multiple entities (`Cu0.02Ti0.94Nb2.04O7' and `TiNb2O7') are present, the system sometimes fails to capture all the entities and make the incorrect associations between the entities and the relations. 

A set of examples is shown in tables 
~\ref{tab:op-single-paper} and ~\ref{tab:op-different-relations} which compare the expected ground truth output with the output predicted by the PNM (MatBERT) model. We observe that differences occur is cases of same sentence having multiple relation triplets. 

Table \ref{tab:op-single-paper} compares the entity-relation triplets for a single manuscript (\citet{yu2016atomic}) from the annotated dataset where the system is able to extract multiple relations and the corresponding entities correctly in most of the cases.  Table \ref{tab:op-different-relations} compares the entity-relation triplets for all the five relations (voltage, conductivity, coulombic efficiency, capacity, energy).  

One limitation of our approach is the presence of errors in the distantly supervised training data. But generally some amount of noise is acceptable in any distantly supervised dataset \cite{mintz2009distant} as it saves lot of time and cost over manually annotating large training set - which is also one of the primary aims of this work. Since, our test data is manually annotated so such errors won’t be there in the test set and evaluation results.

